\documentclass{article}
\pdfoutput=1

    \usepackage[preprint]{neurips_2025}

\usepackage[utf8]{inputenc} 
\usepackage[T1]{fontenc}    
\usepackage{hyperref}       
\usepackage{url}            
\usepackage{booktabs}       
\usepackage{amsfonts}       
\usepackage{nicefrac}       
\usepackage{microtype}      
\usepackage{xcolor}         

\usepackage{color}
\usepackage{algorithm}
\usepackage{algorithmic}
\usepackage{graphicx}
\usepackage{amsmath}
\usepackage{amsfonts}
\usepackage{amssymb}
\usepackage{amsthm}
\usepackage{mathtools}
\usepackage{multirow}
\usepackage{balance} 
\usepackage{mathrsfs}
\usepackage{makecell}
\usepackage{booktabs,array}

\usepackage{natbib}
\setcitestyle{numbers,square}
\setcitestyle{square}

\newtheorem{theorem}{Theorem}

\newtheorem{corollary}[theorem]{Corollary}

\newcommand{\ModelName}{{HDC}}

\title{Making Language Model a \\Hierarchical Classifier}
\author{}

\author{%
  \textbf{Yihong Wang\textsuperscript{1}, Zhonglin Jiang\textsuperscript{1}, Ningyuan Xi\textsuperscript{1}, Yue Zhao\textsuperscript{1}, Qingqing Gu\textsuperscript{1}, Xiyuan Chen\textsuperscript{1},} \\
  \textbf{Hao Wu\textsuperscript{2}, Sheng Xu\textsuperscript{1}, Hange Zhou\textsuperscript{1}, Yong Chen\textsuperscript{1}, Luo Ji\textsuperscript{1*}}\thanks{Corresponding author.} \\
 \textsuperscript{1}Geely AI Lab, \textsuperscript{2}Tsinghua University \\
  \texttt{\{Yihong.Wang2, Luo.Ji1\}@geely.com}
}

\begin{document}

\maketitle

\begin{abstract}

Decoder-only language models, such as GPT and LLaMA, generally decode on the last layer. Motivated by human's hierarchical thinking capability, we propose that a hierarchical decoder architecture could be built with different layers decoding texts simultaneously. Due to limited time and computationally resources, we choose to adapt a pretrained language model into this form of hierarchical decoder. Language heads of the last layer are copied to different selected intermediate layers, and fine-tuned with different task inputs. By thorough experiments, we validate that these selective intermediate layers could be adapted to speak meaningful and reasonable contents, and this paradigm of hierarchical decoder can obtain state-of-the-art performances on multiple tasks such as hierarchical text classification, classification-guided generation, and hierarchical text generation. HdLM outperforms all baselines on WoS, DBpedia, ESconv, EmpatheticDialogues, and several cognitive tests. We also provide thorough theoretical analysis to validate the convergence and computational savings of our methodology. This study suggests the possibility of a generalized hierarchical reasoner, pretraining from scratch.

\end{abstract}

\section{Introduction}
\label{sec:introduction}


Modern Large language models (LLM), such as GPT \cite{openai2024gpt4technicalreport} and Llama \cite{llama3modelcard}, have made remarkable progresses on natural language tasks  \cite{naveed2024comprehensiveoverviewlargelanguage}. Based on the decoder-only architecture, such language models exhibit impressive generalizability and scalability on different tasks and scenarios. Among these tasks, textual reasoning is always challenged which often requires the model to explicitly plan the immediate steps and bootstrap the long-term rewards. To enhance such capabilities, chain-of-thought (CoT) \cite{wei2022COT} and corresponding finetuned or test-time scaling methods \cite{zelikman2022Star,zelikman2024QuietStar,hao2024COCONUT} are proposed, with the thought formulated either on the token-space or the latent-space.

However, these reasoning LLMs do not have an explicit hierarchical thinking mechanism, therefore might suffer from suboptimal solutions on complex reasoning tasks \cite{lcmteam2024largeconceptmodelslanguage}. Their reasoning capabilities are constrained from the scale of annotated data, and the test-time scaling might be upper-bounded \cite{yue2025doesreinforcementlearningreally}. On the other hand, mankind is naturally empowered with hierarchical thinking capabilities \cite{lcmteam2024largeconceptmodelslanguage}, typically in two aspects:

\noindent (i) Different abstraction of conceptual levels: the coarse-grained, strategic decisions guide the generation of fine-grained, detailed decisions.

\noindent (ii) Sequential determination: a key determination is first selected from available options while subsequent options are subject to the selection.

Motivated from these hierarchical thinking mechanisms, models with automatic hierarchical thinking should be explored. Similar studies were mainly based on encoder-decoder architectures and may be classified into model-wise, scale-wise and layer-wise prototypes \cite{liang2019cascadedecoderuniversaldecoding}.

In this paper, we propose a new type of decoder called Hierarchical Decoding Classifier ({\ModelName}), combining the model-wise and layer-wise prototypes. We inherit the strong semantic understanding and generation capabilities of modern LLM, but allow its multiple layers to decode plausible textual contents. An earlier layer may produce more prerequisite, coarse-grained expressions which also work for the context of resulted, fine-grained expressions, which are decoded by later layers. Such type a language model can automatically produce reasonable performance on sequential pipelines of classification and generation tasks, which are verified by our substantial experiments. Comparing with the conventional reasoning LLM, {\ModelName} also enjoys computational savings for both training and inference, which are both verified by theoretical derivation and empirical observations. Our main contributions can be summarized as follows:\\
\noindent (1) We propose {\ModelName}, which can deal with tasks of hierarchical text classification.\\
\noindent (2) We provide a theoretical analysis on the computational benefit {\ModelName}, and also a discussion on its convergence. \\
\noindent (3) We conduct substantial experiments and corresponding analysis to verify {\ModelName}'s performance, as well as its out-of-domain generalization capability.\\
\noindent (4) We discuss about its connection with alternative solutions such as vanilla SFT, cascade, multi-hop, and its scalability and the possibility of an end-to-end framework.

\section{Method}
\label{sec:method}

In this section, we first formalize the problem, then propose a dual-layer fine-tuning mechanism, and finally a two-pass inference paradigm.

\subsection{Problem Formulation} 


Given a language model i$\mathcal{M}$, its total number of layers is $K$, and the language head is $\mathcal{H}$. Different from traditional query-response tasks, here we try to define a generalized hierarchical textual tasks. Given a user query $q$, hierarchical textual tasks requires the agent to generate a sequence of responses $\mathbf{r}_{1:D} := \{ r_1, r_2, \cdots, r_D \}$, in which $D$ is the hierarchical depth. The standard format of hierarchical textual data then becomes $(q, \mathbf{r}_{1:D})$. We use $L$ and $L_d$ to denote the lengths of $q$ and $r_d$, respectively.

The above hierarchical textual task can be solved recursively. That is, for each step $i \in \{1, 2, \cdots, D\}$, the current response can be produced grounded by the query and prior responses:
\begin{equation}
    \mathcal{T}_i \in \{ \mathcal{C}, \mathcal{G} \}: r_i \leftarrow \mathcal{M}(q, \mathbf{r}_{1:i-1})
\end{equation}
where $\mathcal{T}$ denotes a uni-step subtask. Here we further argue $\mathcal{T}$ can be generally classified into two categories: the classification task $\mathcal{C}$ and the generation task $\mathcal{G}$. As indicated in Figure \ref{fig:task}, the following different paradigms can be summarized from hierarchical textual tasks:




\paragraph{Hierarchical text classification (HTC):}classifications are asked from the coarse-grained label to the fine-grained label. The lower-level label candidates are usually constrained by the choice of the higher-level label. We denote this paradigm as $\mathcal{C} \rightarrow \mathcal{C}$. 

While there are evident methods to solve hierarchical textual tasks, either by computation time scaling (such as CoT or reasoning LLM), or the multi-hop inference, in this paper we propose a single-model framework which has less computational overheads.


\subsection{Architecture} 

In this section, we propose an adjusted architecture based on the decoder-only transformer, which can also deal with hierarchical textual tasks by post-hoc adaptations. We first make a reasonable assumption that the depth of hierarchical textual task is smaller, \textit{i.e.}, $D < K$. Then we select $D-1$ intermediate layers with their indices satisfying:
\begin{equation}
    k \in \{ k_1, \cdots, k_{D-1}, K \}, 0 < k \leq K
\end{equation}
These layers, along with the final layer, are used to decode $D$ responses. To achieve this objective, we replicate the language heads in the $K$-th layer to the $D-1$ intermediate layers, with the parameter randomly initialized.
\begin{equation}
    \mathcal{H}_d \leftarrow \mathcal{H}_K, d = 1, 2, \cdots, D-1
\end{equation}

To make the derivation clearer, we propose some notations. Given a language model $\mathcal{M}$, $\mathcal{M}_k$ represents its $k$-th layer. The forward pass from the $k_1$ to $k_2$-th layer is denoted by $\mathcal{M}_{k_1:k_2}$, while $\mathcal{M}_{k_1:k_2}^{\xrightarrow{L_1}}$ denotes the forward pass plus a textual generation.

The final layer still holds the  standard supervised Fine-Tuning (FT) loss:
\begin{equation}
    \mathcal{L}^{\text{FT}}_D = - \frac{1}{L_D} \sum_{j=1}^{L_D} \log \left[ \text{P}(r_{D}(j)|q, \mathbf{r_{1:D-1}}, r_{D}(1:j-1)) \right]
    \label{eq:sft}
\end{equation}
where $j$ denotes the $j$-th token of $r_D$. 

\subsection{Training} 

The $D-1$ intermediate layers can not decode reasonably with newly added heads. To educate them to decode proficient language, post-hoc adaptation is needed. 


Given the hierarchical textual sample $(q, r_{0:d})$, the latent vector of each hierarchical level can be calculated recursively by the forward pass of each LM block: 
\begin{align}
    e_1 &= \mathcal{M}_{0:k_1}(q), \quad e_{d+1} = \mathcal{M}_{k_d:k_{d+1}}(e_d, r_d), \quad \mathcal{L}_d = \mathcal{L}^{\text{FT}}(q, e_{1:d}, r_{1:d}), \quad d = 1, 2, \cdots, D
\end{align}
with the finetuning loss $\mathcal{L}_d$ implemented on each decoding layer. The final loss can then be the linear combination of them: $\mathcal{L} = \sum_{d=1}^{D-1} f_d \mathcal{L}_{d} + \mathcal{L}_{D}$
with the list of $\{ f_1, \cdots, f_{D-1}, 1  \}$ as the loss weights. During the training, we implement the causal mask for both prior tokens and subsequent responses. 


\subsection{Inference} 

We keep a similar forward logic of inference to the training. For the query $q$, we still calculate its logits throughout all attention layers, similar with the conventional language model. For each $k$-th layer, the prior latent vector is employed to decode the $k$-th response, while the new decoded latent is also employed to forward pass the final layer, until the final response is decoded grounded by all previous latents. 
\begin{align}
    e_1, \hat{L}_1 &= \mathcal{M}_{0:k_1}^{\xrightarrow{L_1}}(q), \quad e_{d+1}, \hat{L}_{d+1} = \mathcal{M}_{k_d:k_{d+1}}^{\xrightarrow{L_d}}(e_d) \\ \hat{r}_d &= \mathcal{H}_{k_d}(e_d(L+\sum{\hat{L}_{<d-1}}:)), \quad d = 1, 2, \cdots, D
\end{align}
Note the above expression has a similar structure to the state-space model (SSM).

\section{Experiment}
\label{sec:experiment}



\subsection{Datasets}



\paragraph{HTC ($\mathcal{C} \rightarrow \mathcal{C}$).} We use the famous WoS \cite{toit2024introducingnewbenchmarkdatasets} (with depth $D=2$) and DBpedia \cite{10.1007/978-3-540-76298-0_52} (with depth $D=3$) as the training datasets. WoS are abstracts of published papers from Web of Science while DBpedia extracts structured information from Wikipedia.



\subsection{Setting}

\paragraph{Implementation Details.} We conduct a post-hoc adaptation on the basis of Llama3-8B-Instruct \cite{grattafiori2024llama3herdmodels}, which has a total $K = 32$ attention layers. During training, we use the AdamW optimizer with decay of 0.01 and the cosine scheduler. The training batch size is 16 and the sequence length is 2048. The experiment is running on LlamaFactory \cite{zheng2024llamafactory} with 32 A100 GPUs, lasting about 16 hours. Other settings, which is dataset-wise, are listed in the Appendix.


\paragraph{Classification metrics.} F1-related scores including Micro-F1 and Macro-F1 are used. Micro-F1 considers the overall precision and recall of all instances, while Macro-F1 equals the average F1-score of labels. For CgG tasks, we also provide the classification accuracy, and the preference $bias$ as defined by \cite{kang-etal-2024-large} based on Bradley-Terry model~\citep{bradley1952btmodel} \footnote{Detailed formula in the Appendix. Smaller $bias$ means better.}. For HTC tasks, we provide the metric results on the bottom level of labels.



\subsection{Results}

\paragraph{Hierarchical Textual Classification.} Table \ref{tab:htc_result} shows the F1 results of WoS (depth=2) and DBpedia (depth=3), compared to previous state-of-the-art HTC baselines. Our {\ModelName} performs the best on both WoS and DBpedia, indicating it has a reasonable semantic hierarchy comprehension and conducts accurate bottom-level classification. For typical cases of {\ModelName} on HTC, see Appendix.

\begin{table*}[h!]
\centering
\caption{F1 scores of hierarchical textual classification on WoS and DBPedia.}
\label{tab:htc_result}
\small
\begin{tabular}{ll|cc|cc}
\toprule
 & \multicolumn{1}{c|}{Dataset($\rightarrow$)} & \multicolumn{2}{c|}{WoS ($\mathcal{C} \rightarrow \mathcal{C}$)} & \multicolumn{2}{c}{DBPedia ($\mathcal{C} \rightarrow \mathcal{C} \rightarrow \mathcal{C}$)}  \\
\cline{2-2}  \cline{3-4} \cline{5-6}
 & Method ($\downarrow$) & Micro-F1 & Macro-F1 & Micro-F1 & Macro-F1 \\
\toprule
\parbox[t]{5mm}{\multirow{3}{*}{\rotatebox[origin=c]{90}{
\parbox[c]{1.2cm}{\centering \scriptsize Retrieval-based}}}}
& HierVerb \cite{ji-etal-2023-hierarchical} & 80.93 & 73.80 & 96.17 & 93.28 \\
& Retrieval \cite{chen-etal-2024-retrieval-style} & 81.12 & 73.72 & 96.22 & 93.37 \\
& Retrieval-ICL \cite{chen-etal-2024-retrieval-style} & 78.62 & 69.56 & 95.56 & 92.04 \\
\midrule
\parbox[t]{5mm}{\multirow{6}{*}{\rotatebox[origin=c]{90}{
\parbox[c]{1.2cm}{\centering \scriptsize Bert-based}}}}
& BERT & 86.28 & 80.58 & 95.31 & 89.16 \\
& \; + HiAGM \cite{zhou-etal-2020-hierarchy} & 86.04 & 80.19 & - & - \\
& \; + HiMatch \cite{chen-etal-2021-hierarchy} & 86.70 & 81.06 & - & - \\
& \; + softprompt \cite{wang-etal-2022-hpt} & 86.57 & 80.75 & - & - \\
& HGCLR \cite{wang-etal-2022-incorporating} & 87.11 & 81.20 & 95.49 & 89.41 \\
& HPT \cite{wang-etal-2022-hpt} & 87.16 & 81.93 & 96.13 & 93.34 \\
\cline{1-6}
 & \textbf{{\ModelName}} (ours) & \bf 88.40 & \bf 87.54 & \bf 96.73 & \bf 96.37 \\
\bottomrule
\end{tabular}
\end{table*}

\paragraph{Ablation on methodology components.} We conduct the following ablation studies: 

\noindent (a) \textit{w/ head}: implement the classification head on the intermediate layers and the final layer, and use them to study the classification tasks.

\noindent (b) \textit{cascade}: use two separate LLMs, finetuned and inferenced by $r_1$ and $r_2$, respectively.

\noindent (c) \textit{multi-hop}: inference a mix-finetuned LLM twice, with $r_1$ and $r_2$ inferenced sequentially.

\noindent (d) \textit{sft@$K$}: use the final layer to decode both $r_1$ and $r_2$.

Table \ref{tab:ablation_wos} shows the ablation result on WoS. {\ModelName} still holds the best results which indicates our framework is rigorous. Furthermore, we also compare {\ModelName} to two imaginary solutions 

\noindent (e) \textit{true-1}: use the ground truth $r_1$ and only conduct the $r_2$ classification (from the full $r_2$ list).

\noindent (f) \textit{true-1 \& true-2 candidates}: use the ground truth $r_1$ as well as ground truth  $r_2$ candidates.

both of which are apparently not applicable in the practice. Even so, {\ModelName}'s performance is still close to them, indicating a high-level comprehension of label hierarchy.

\begin{table}[t!]
\caption{Ablation on WoS (HTC task). miF1 respresents micro-F1 and maF1 represents macro-F1.}
\label{tab:ablation_wos}
\centering
\small
\begin{tabular}{c | ccccc | cc}
    \toprule
    \multicolumn{1}{c|}{\multirow{2}[2]{*}{Metrics}} &  \multicolumn{5}{c|}{Ablation} &  \multicolumn{2}{c}{\textit{\textcolor{gray}{Not applicable practically}}} \\
    \cline{2-6} \cline{7-8}
     & w/ head & cascade & multi-hop & sft@$K$ & \textbf{\ModelName} & \textcolor{gray}{true-1} & \textcolor{gray}{true-1 \& true-2 candidates} \\ %
    \midrule
    Micro-F1 & 82.38 & 87.92 & 64.51 & 86.64 & \bf 88.40 & \textcolor{gray}{89.44} & \textcolor{gray}{92.78} \\
    Macro-F1 & 37.24 & 86.84 & 68.62 & 85.86 & \bf 88.54 & \textcolor{gray}{88.61} & \textcolor{gray}{89.21} \\ 
    \bottomrule
\end{tabular}
\end{table}

\paragraph{Computational time.} We also calculate the computational times in the experimental environment. For different values of $L_2$ (128, 256, 512), the computation reduction of {\ModelName} grows linearly (7.03\%, 18.03\%, 26.25\%), no matter what the input length is. On the other hand, for different $k$ (16, 20, 24), the computation reduction of {\ModelName} also grows linearly (7.43\%, 12.09\%, 17.85\%). These observations verify the theoretical conclusion. Detailed plots and corresponding error bars are included in the Appendix.



\section{Related Work}

\paragraph{Large language models with latent reasoning pace.} Token-level LLMs are repurposed to reasoning on the latent space to further enhance the thinking capability. Recently, COCONUT \citep{hao2024COCONUT} utilizes the latent state of the LLM to represent the reasoning state, which forms a continuous thought. TRICE \cite{hoffman2023training} samples and finetunes the chained rationales by Monte-Carlo sampling.  LCM \citep{lcmteam2024largeconceptmodelslanguage} studies the sentence-level conceptual embeddings. However, these studies generally reason and decode the latent thoughts on the final layer, which lacks of a structured hierarchical view; while our method constructs a recursive chain of layer latent which is a natural hierarchal thinker.


\paragraph{Hierarchical Decoding.} There were early attempts at hierarchical decoding mechanism. Cascade decoder \cite{liang2019cascadedecoderuniversaldecoding} employs a cascade branching structure on the biomedical image segmentation tasks. CoHD \cite{luo2024cohdcountingawarehierarchicaldecoding} proposes a counting-aware hierarchical decoding framework for image segmentation. Su et.al \cite{su-etal-2018-natural} introduces a hierarchical decoding NLG model based on different levels of linguistic patterns. ExHiRD \cite{chen-etal-2020-exclusive} designs a hierarchical decoding encoder-decoder structure, dividing the keyphrase generation task into phase-level decoding and word-level decoding. HSD \cite{zhu2024hierarchicalskipdecodingefficient} adaptively skips decoding layers in a hierarchical manner. In contrast, we propose a model-based framework based on LLM, covering both training and inference.

\section{Conclusion}

In this study, we propose a hierarchical decoding language model called {\ModelName}, which can provide both sequential and strategic understanding and generation capabilities. Post-hoc adapted from pretrained language models, {\ModelName} can achieve state-of-the-art performance on hierarchical text classification, classification-guided generation, and hierarchical text generation. We also conduct theoretical analysis on its computational efficiency and convergence. {\ModelName} sheds some lights on a generalized, automatically hierarchical artificial thinker.

\bibliographystyle{nips}
\bibliography{main}

\clearpage

\end{document}